\documentclass[10pt,twocolumn,letterpaper]{article}

\usepackage{cvpr}              

\usepackage[table]{xcolor}
\usepackage{times}
\usepackage{epsfig}
\usepackage{graphicx}
\usepackage{amsmath}
\usepackage{amssymb}
\usepackage{booktabs}
\usepackage{multirow}
\usepackage{multicol}
\usepackage{caption}
\usepackage{makecell}
\usepackage{float}
\usepackage{pifont}
\usepackage[accsupp]{axessibility}  

\newcommand{\cmark}{\ding{51}}%
\newcommand{\bK}{\mathbf{K}}
\newcommand{\bR}{\mathbf{R}}

\usepackage[pagebackref,breaklinks,colorlinks]{hyperref}

\usepackage[capitalize]{cleveref}
\crefname{section}{Sec.}{Secs.}
\Crefname{section}{Section}{Sections}
\Crefname{table}{Table}{Tables}
\crefname{table}{Tab.}{Tabs.}

\def\cvprPaperID{2307} 
\def\confName{CVPR}
\def\confYear{2023}

\begin{document}

\title{Shape-Constraint Recurrent Flow for 6D Object Pose Estimation}


\author{%
	{Yang Hai $^1$, \quad Rui Song $^1$, \quad Jiaojiao Li $^1$, \quad Yinlin Hu $^2$} \\
	{\small $^1$ State Key Laboratory of ISN, Xidian University, \quad $^2$ MagicLeap} \\
}

\maketitle

\begin{abstract}

Most recent 6D object pose methods use 2D optical flow to refine their results. However, the general optical flow methods typically do not consider the target's 3D shape information during matching, making them less effective in 6D object pose estimation. In this work, we propose a shape-constraint recurrent matching framework for 6D object pose estimation. We first compute a pose-induced flow based on the displacement of 2D reprojection between the initial pose and the currently estimated pose, which embeds the target's 3D shape implicitly. Then we use this pose-induced flow to construct the correlation map for the following matching iterations, which reduces the matching space significantly and is much easier to learn. Furthermore, we use networks to learn the object pose based on the current estimated flow, which facilitates the computation of the pose-induced flow for the next iteration and yields an end-to-end system for object pose. Finally, we optimize the optical flow and object pose simultaneously in a recurrent manner. We evaluate our method on three challenging 6D object pose datasets and show that it outperforms the state of the art significantly in both accuracy and efficiency.

\end{abstract}

\section{Introduction}

6D object pose estimation, {\it i.e.}, estimating the 3D rotation and 3D translation of a target object with respect to the camera, is a fundamental problem in 3D computer vision and also a crucial component in many applications, including robotic manipulation~\cite{collet2011moped} and augmented reality~\cite{marchand2015pose}.
Most recent methods rely on pose refinement to obtain accurate pose results~\cite{RNNPose_2022_cvpr, PFA, coupled_2022_cvpr}. 
Typically, they first synthesize an image based on the rendering techniques~\cite{BlenderProc_2019_arxiv, Pytorch3d_2020_arxiv} according to the initial pose, then estimate dense 2D-to-2D correspondence between the rendered image and the input based on optical flow networks~\cite{RAFT_2020_eccv}. After lifting the estimated 2D optical flow to 3D-to-2D correspondence based on the target's 3D shape, they can obtain a new refined pose using Perspective-n-Points (PnP) solvers~\cite{EPnP_2009_ijcv}.

\begin{figure}[t]
   \centering
   \setlength\tabcolsep{1pt}
   \begin{tabular}{cccc}
   \includegraphics[width=0.24\linewidth]{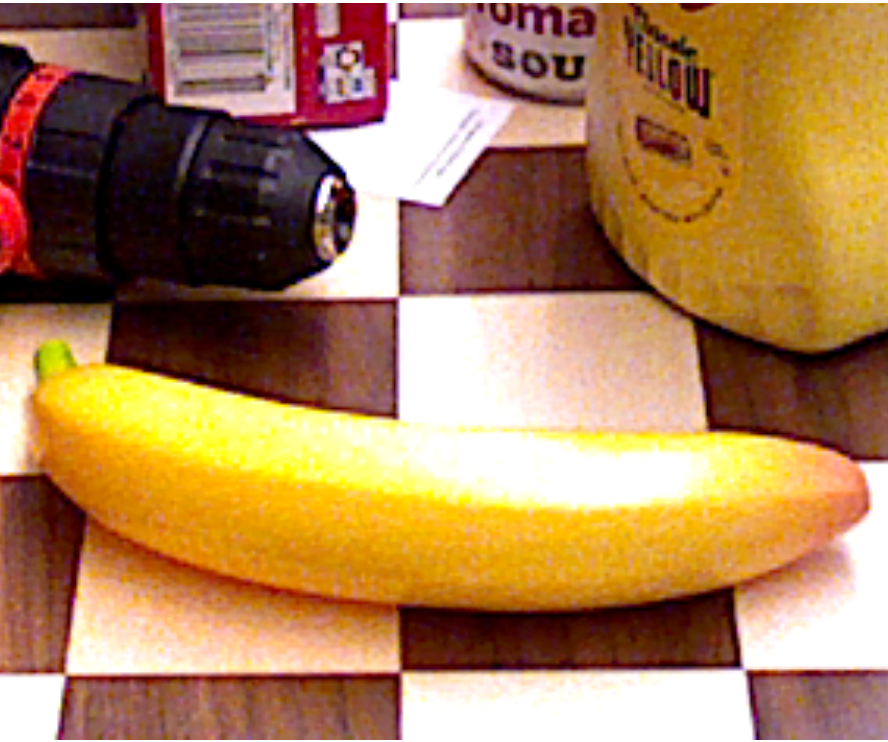} &
   \includegraphics[width=0.24\linewidth]{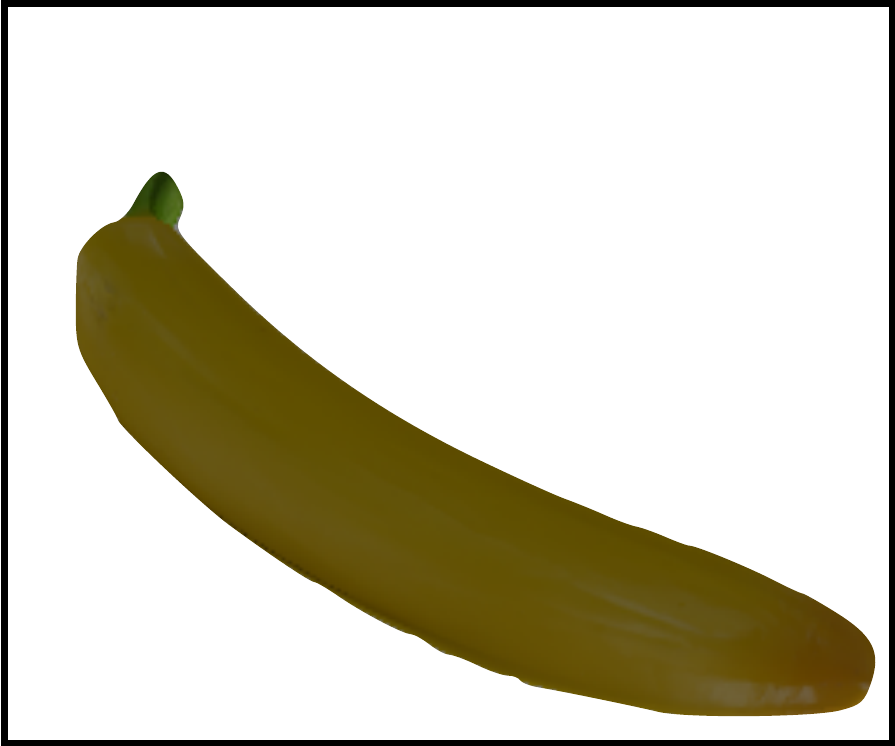} &
   \includegraphics[width=0.24\linewidth]{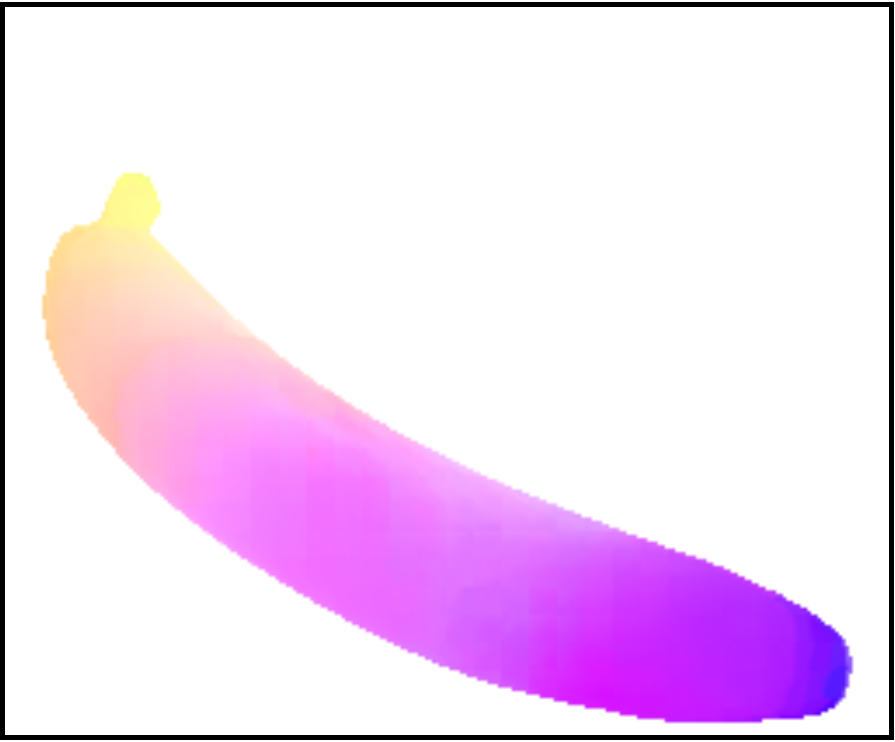} &
   \includegraphics[width=0.24\linewidth]{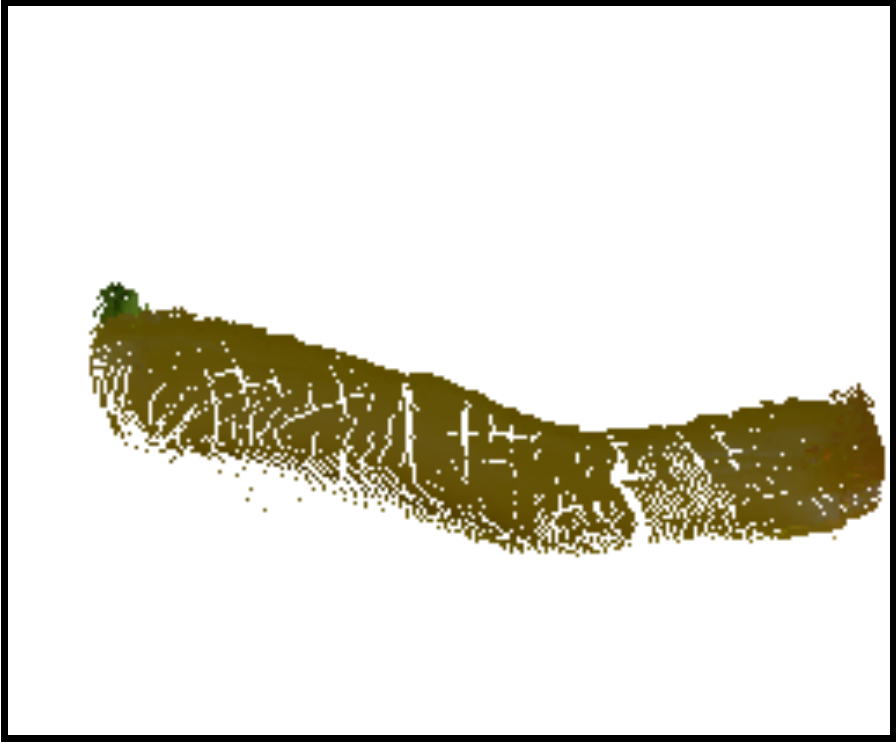} \\
   \small{Input} & \small{Initialization} & \small{Flow} & \small{Flow warp} \\
   \end{tabular}
   \caption{\textbf{The problem of optical flow in 6D pose estimation.}
   Given an initial pose, one can estimate the dense 2D-to-2D correspondence (optical flow) between the input and the synthetic image rendered from the initial pose, and then lift the dense 2D matching to 3D-to-2D correspondence to obtain a new refined pose by PnP solvers (PFA-Pose~\cite{PFA}). However, the flow estimation does not take the target's 3D shape into account, as illustrated by the warped image based on the estimated flow in the last figure, which introduces significant matching noise to pose solvers and is suboptimal for 6D object pose estimation.
   }
   \label{fig:teaser}
\end{figure}

Although this paradigm works well in general, it suffers from several weaknesses.
First, the general optical flow networks they use are mainly built on top of two assumptions, {\it i.e.}, the brightness consistency between two potential matches and the smoothness of matches within a local neighbor~\cite{HSFlow}. These assumptions, however, are too general and do not have any clue about the target's 3D shape in the context of 6D object pose estimation, making the potential matching space of every pixel unnecessarily large in the target image. Second, the missing shape information during matching often results in flow results that do not respect the target shape, which introduces significant matching noise, as shown in Fig.~\ref{fig:teaser}. Third, this multi-stage paradigm trains a network that relies on a surrogate matching loss that does not directly reflect the final 6D pose estimation task~\cite{Single-stage_2020_cvpr}, which is not end-to-end trainable and suboptimal.

\begin{figure*}[t]
    \centering
    \setlength\tabcolsep{15pt}
    \begin{tabular}{c@{\hspace{2em}}c}
        \includegraphics[width=0.45\linewidth]{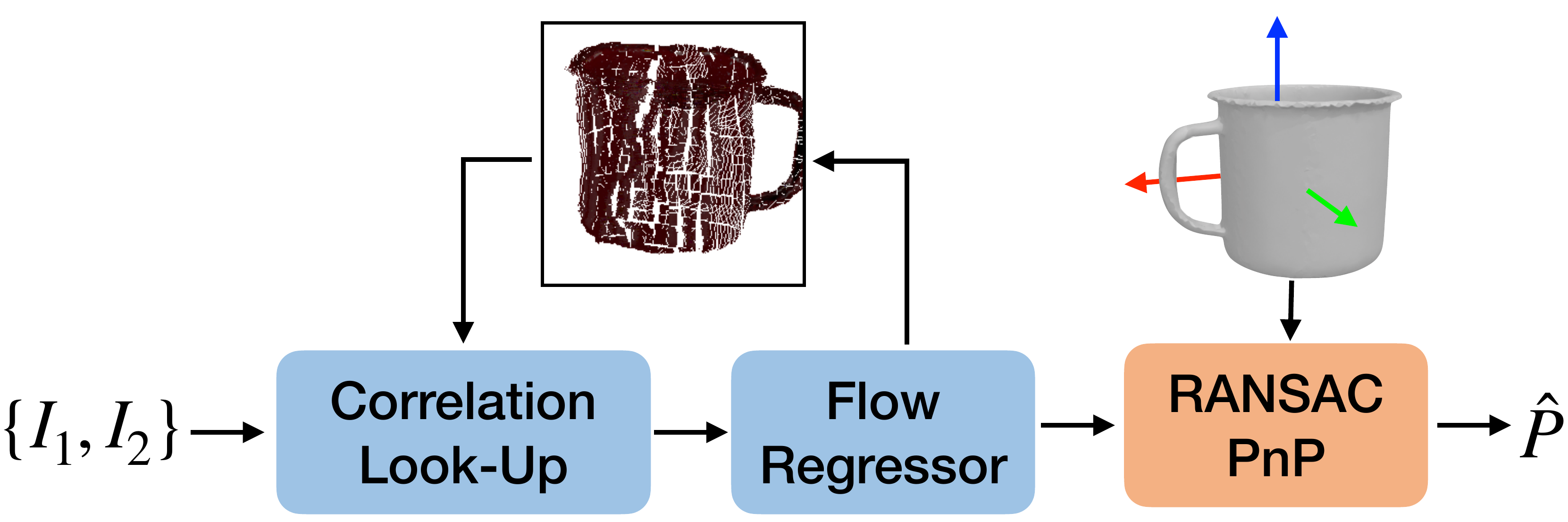} &
        \includegraphics[width=0.45\linewidth]{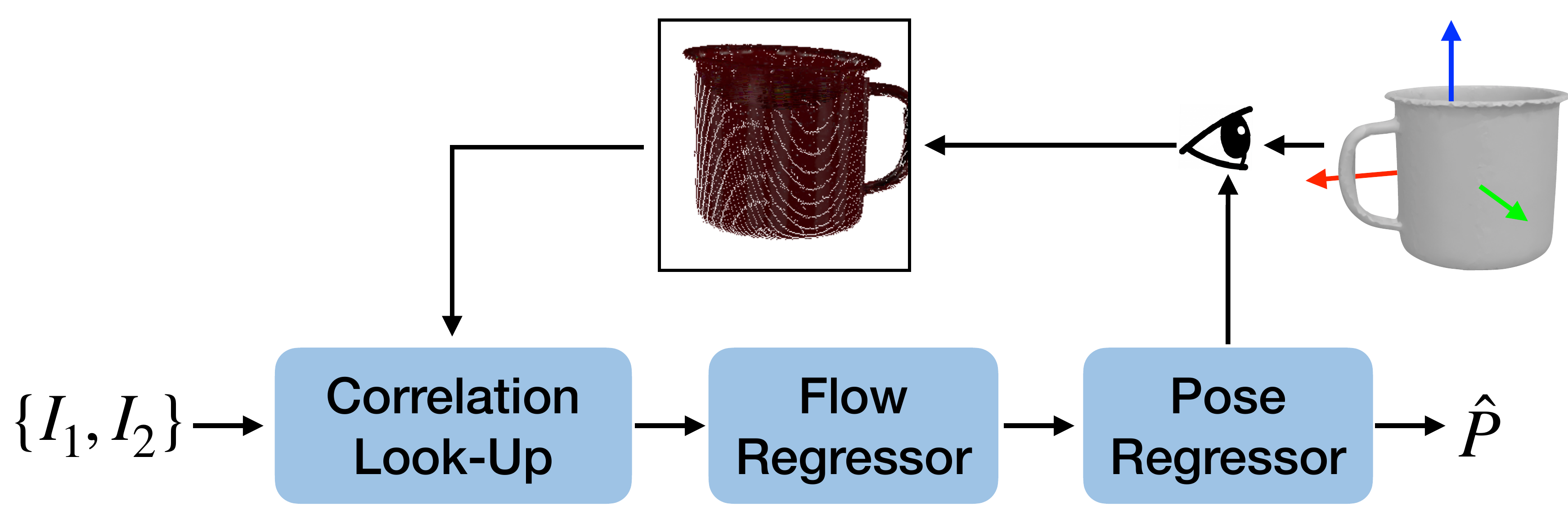} \\
        \small{(a) The standard strategy} & \small{(b) Our strategy} \\
    \end{tabular}
    \caption{{\bf Different pose refinement paradigms.}
    {\bf (a)} Most pose refinement methods~\cite{PFA} rely on a recurrent architecture to estimate dense 2D flow between the rendered image $I_1$ and the real input image $I_2$, based on a dynamically-constructed correlation map according to the flow results of the previous iteration. After the convergence of the flow network and lifting the 2D flow to a 3D-to-2D correspondence field, they use PnP solvers to compute a new refined pose $\hat{P}$. This strategy, however, has a large matching space for every pixel in constructing correlation maps, and optimizes a surrogate matching loss that does not directly reflect the final 6D pose estimation task. {\bf (b)} By contrast, we propose optimizing the pose and flow simultaneously in an end-to-end recurrent framework with the guidance of the target's 3D shape. We impose a shape constraint on the correlation map construction by forcing the construction to comply with the target's 3D shape, which reduces the matching space significantly. Furthermore, we propose learning the object pose based on the current flow prediction, which, in turn, helps the flow prediction and yields an end-to-end system for object pose.
    }
    \label{fig:teaser_compare}
\end{figure*}

To address these problems, we propose a shape-constraint recurrent matching framework for 6D object pose estimation. It is built on top of the intuition that, in addition to the brightness consistency and smoothness constraint in classical optical flow solutions~\cite{Sintel_2012_eccv, Flyingthings3D-2016_cvpr}, the dense 2D matching should comply with the 3D shape of the target.  
We first build a 4D correlation volume between every pixel of the source image and every pixel of the target image, similar to RAFT~\cite{RAFT_2020_eccv}. While, instead of indexing from the correlation volume according to the current flow during the iteration, we propose indexing the correlation volume based on a pose-induced flow, which is forced to contain only all the 2D reprojections of the target's 3D shape and reduces the matching space of the correlation map construction significantly. Furthermore, we propose to use networks to learn the object pose based on the current flow prediction, which facilitates the computation of
the pose-induced flow for the next iteration and also removes the necessity of explicit PnP solvers, making our system end-to-end trainable and more efficient, as shown in Fig.~\ref{fig:teaser_compare}(b).

We evaluate our method on the challenging 6D object pose benchmarks, including LINEMOD~\cite{Linemod_2012_accv}, LINEMOD-Occluded~\cite{OccLinemod_2015_iccv}, and YCB-V~\cite{PoseCNN_2018_rss}, and show that our method outperforms the state of the art significantly, and converges much more quickly.

\section{Related Work}

\noindent\textbf{Object pose estimation}, has shown significant improvement~\cite{PoseCNN_2018_rss, PVNet_2019_cvpr,Densefusion_2019_cvpr} after the utilization of deep learning techniques~\cite{resnet_2016_cvpr, resnext_2017_cvpr}. While most of them still follow the traditional paradigm, which consists of the establishment of 3D-to-2D correspondence and the PnP solvers. Most recent methods create the correspondence either by predicting 2D points of some predefined 3D points~\cite{Bb8_2017_iccv,Segdriven_2019_cvpr,PVNet_2019_cvpr,WDR_2021_cvpr} or predicting the corresponding 3D point for every 2D pixel location within a segmentation mask~\cite{DPOD_2019_ICCV,ZebraPose_2022_cvpr,CDPN_2019_iccv,Gdr-net_2021_cvpr,So-Pose_2022_iccv,RLLG_2020_cvpr}.
On the other hand, some recent methods try to make the PnP solvers differentiable~\cite{Single-stage_2020_cvpr, BPnP_2020_cvpr, EroPnP}.
However, the accuracy of these methods still suffers in practice. We use pose refinement to obtain more accurate results in this work.

\noindent\textbf{Object pose refinement}, usually relies on additional depth images~\cite{AAE_2018_eccv,PoseCNN_2018_rss,CDPN_2019_iccv,Densefusion_2019_cvpr}, which is accurate but the depth images are hard to obtain in some scenarios and even inaccessible in many applications~\cite{WDR_2021_cvpr, WaterPose}. Most recent refinement methods use a render-and-compare strategy without any access to depth images and achieve comparable performance~\cite{Deep-model_2018_eccv,DeepIM_2018_eccv,DPOD_2019_ICCV,cosypose_2020_eccv,Bb8_2017_iccv,PFA,RNNPose_2022_cvpr, Repose_2021_iccv,coupled_2022_cvpr}.
These methods, however, usually formulate pose refinement as a general 2D-to-2D matching problem and do not consider the fact that the dense 2D matching should comply with the 3D shape of the target, which is suboptimal in 6D object pose estimation. On the other hand, most of them rely on numerical PnP solvers~\cite{EPnP_2009_ijcv} as their post processing and optimize a surrogate matching loss that does not directly reflect the final 6D pose estimation task during training.
By contrast, we propose a recurrent matching framework guided by the 3D shape of the target, which transforms the constraint-free matching problem into a shape-constraint matching problem. Furthermore, we propose to learn the object pose from the intermediate matches iteratively, making our method end-to-end trainable and producing more accurate results.

\noindent\textbf{Optical flow estimation}, whose goal is to obtain the matching of every pixel from the source image to the target image, has been widely studied for a long time~\cite{HSFlow}. Classically, it is formulated as an energy optimization problem, which is usually based on the assumption of brightness consistency and local smoothness~\cite{HSFlow, hu2017robust, hu2016efficient, Fullflow_2016_cvpr,EpicFlow}. Recently, the learning-based methods inspired by those traditional intuitions have shown great progress in estimating flow in large-displacement and occlusion scenarios~\cite{HSFlow,Flownet_2015_iccv,Flonet2.0_2017_cvpr,Pwcnet_2018_cvpr,GMA_2021_iccv, craft_2022_cvpr,raft-stereo_2021_3dv}. 
Especially, RAFT~\cite{RAFT_2020_eccv}, which introduces a recurrent deep network architecture for optical flow, has shown significant improvement over the previous learning-based methods further.
However, these optical flow methods are general and can not utilize the target's prior knowledge, making them suboptimal when used in 6D object pose estimation.
By contrast, we propose embedding the target's 3D shape prior knowledge into the optical flow framework for 6D object pose estimation, which reduces the matching space significantly and is much easier to learn.

\begin{figure*}[ht]
    \centering
    \includegraphics[width=1.0\linewidth]{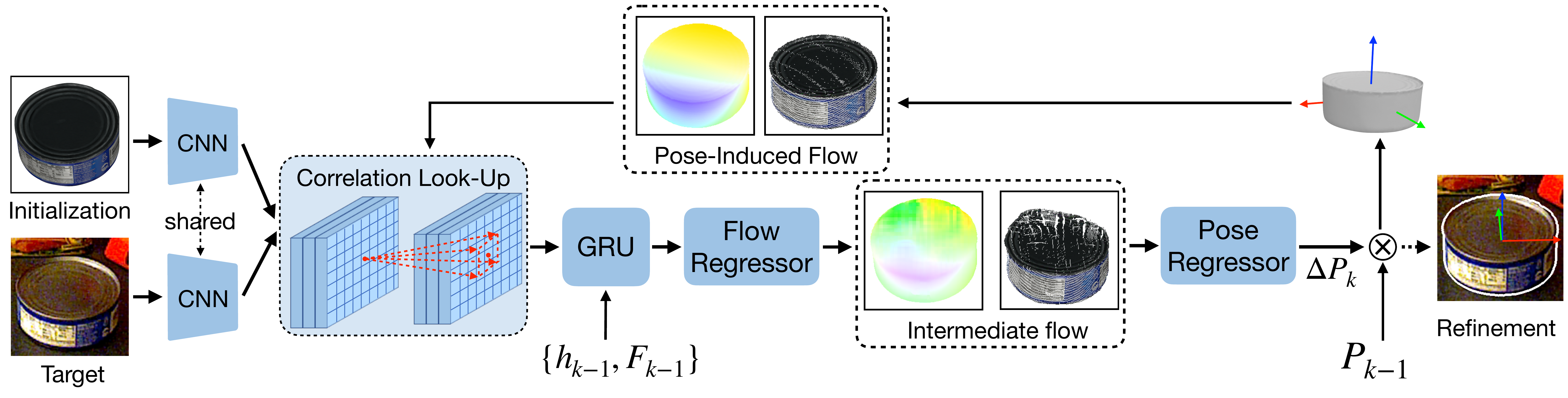}
    \caption{
       \textbf{Overview of our shape-constraint recurrent framework.} 
    After building a 4D correlation volume between the rendered image and the input target image, we use GRU~\cite{GRU} to predict an intermediate flow, based on the predicted flow $F_{k-1}$ and the hidden state $h_{k-1}$ of GRU from the previous iteration. We then use a pose regressor to predict the relative pose $\Delta P_k$ based on the intermediate flow, which is used to update the previous pose estimation $P_{k-1}$. Finally, we compute a pose-induced flow based on the displacement of 2D reprojection between the initial pose and the currently estimated pose $P_{k}$. We use this pose-induced flow to index the correlation map for the following iterations, which reduces the matching space significantly. Here we show the flow and its corresponding warp results in the dashed boxes. Note how the intermediate flow does not preserve the shape of the target, but the pose-induced flow does.
       }
    \label{fig:arch}
\end{figure*}

\section{Approach}
\label{sec:approach}

Given a calibrated RGB input image and the 3D model of the target, our goal is to estimate the target's 3D rotation and 3D translation with respect to the camera. We obtain a pose initialization based on existing pose methods~\cite{WDR_2021_cvpr,PoseCNN_2018_rss}, and refine it using a recurrent matching framework. We focus on the refinement part in this paper. We first have an overview of our framework and then discuss the strategy of reducing the search space of matching by imposing a shape constraint. Finally, we present the design of learning object pose based on optical flow to make our matching framework end-to-end trainable.

\subsection{Overview}
\label{sec:overview}

Given an input image and the initial pose, we synthesize an image by rendering the target according to the initial pose, and use a shared-weight CNN to extract features for both the rendered image and the input image, and then build a 4D correlation volume containing the correlations of all pairs of feature vectors between the two images, similar to PFA~\cite{PFA,RAFT_2020_eccv}. However, unlike the standard strategy that indexes the correlation volume for the next iteration without any constraints, we use a pose-induced flow for the indexing, which embeds the target's shape information implicitly.

We first predict an intermediate flow based on the constructed correlation map. Then we use a pose regressor to predict an intermediate pose based on the intermediate flow. After that, we compute the pose-induced flow based on the displacement of 2D reprojection between the initial pose and the currently estimated pose. We use this pose-induced flow to index the correlation map for the next iteration, which reduces the matching space significantly.
On the other hand, the pose regressor based on the intermediate flow removes the need for RANSAC-PnP and produces an end-to-end system for object pose. Fig.~\ref{fig:arch} shows the overview of our framework.

\subsection{Shape-Constraint Correlation Space}
\label{sec:shape-constraint}

We first obtain a 4D correlation volume $\mathbf{C} \in \mathbb{R}^{H \times W \times H \times W}$ based on the dot product between all pairs of feature vectors from image features from different pyramid levels~\cite{RAFT_2020_eccv}. The standard lookup operation generates a correlation feature map by indexing from the correlation volume, which maps the feature vectors at location $\mathbf{x}=(u, v)$ in $I_1$ to the corresponding new location in $I_2$: $\mathbf{x}' = (u, v) + f(u, v)$, where $f(u, v)$ is the currently estimated flow. This standard lookup operation works well in general, but does not consider the fact that all the matches should comply with the shape of the target in 6D object pose estimation, making its matching space unnecessarily large.

To address this, we embed the target's 3D shape into the lookup operation, generating a shape-constraint location in constructing the new correlation map:
\begin{equation}
\mathbf{x}' = (u_k, v_k) + f(u_k, v_k; {\bf K}, {\bf S}, {\bf P}_0, {\bf P}_k), \; 1 \leq k \leq N,
\end{equation}
where $N$ is the number of iterations, ${\bf K}$ is the intrinsic camera matrix, ${\bf S}$ is the target's 3D shape, ${\bf P}_0$ and ${\bf P}_k$ are the initial pose and the currently estimated pose of the target, respectively. We call the flow fields ${\bf f}(u_k, v_k; {\bf K}, {\bf S}, {\bf P}_0, {\bf P}_k)$ pose-induced flow.

\begin{figure}[t]
    \centering
    \setlength\tabcolsep{10pt}
\begin{tabular}{cc}
  \includegraphics[width=0.35\linewidth]{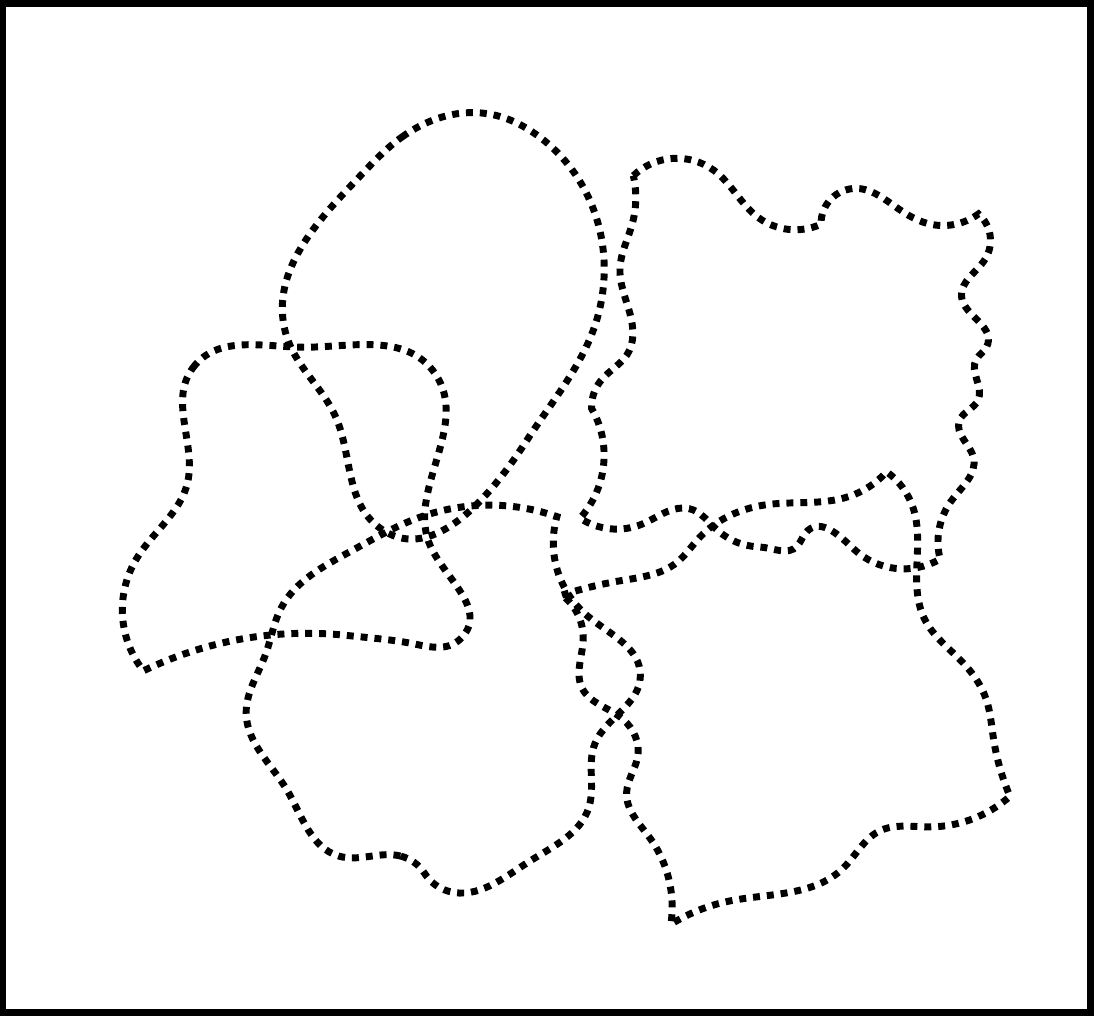} & 
  \includegraphics[width=0.35\linewidth]{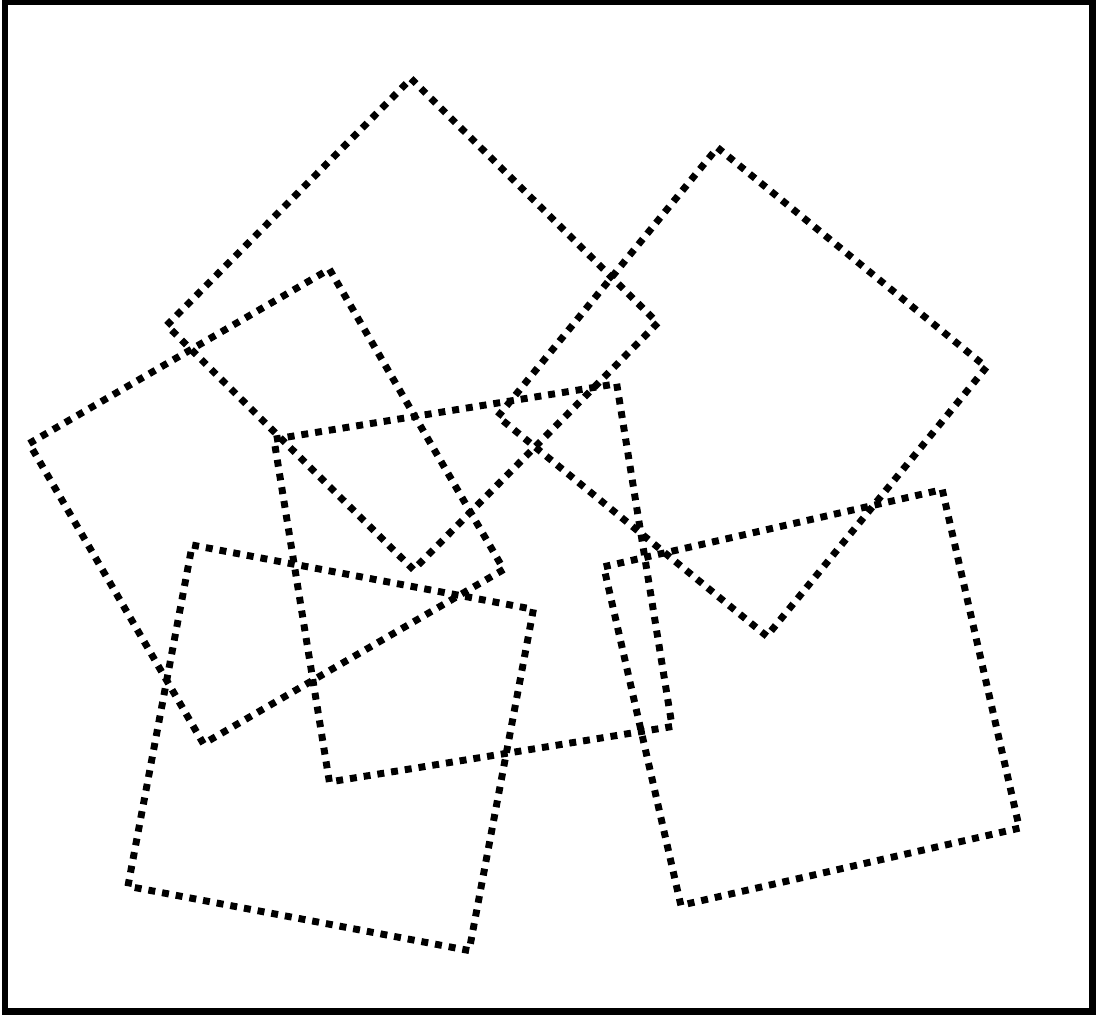} \\
 \small{(a) Standard} & \small{(b) Shape-constraint}
\end{tabular}

    \caption{\textbf{Illustration of shape-constraint correlation space.}
    {\bf (a)} The standard index operation has no constraint in constructing the correlation map, which has unnecessarily large search space for matching in 6D object pose estimation. {\bf (b)} By contrast, we force it to contain only all the 2D reprojections of the target's 3D shape (illustrated as rectangles), reducing the matching space significantly.
   }
   \label{fig:correlation_lookup}
\end{figure}

More specifically, given a 3D point ${\bf p}_i$ on the target's mesh, we use the perspective camera model to get its 2D location ${\bf u}_{i0}$ under the initial pose ${\bf P}_0$,
\begin{equation}
    \begin{aligned}
    \lambda_i
    \begin{bmatrix}
    {\bf u}_{i0} \\
    1 \\
    \end{bmatrix}
    =\bK(\bR_0{\bf p}_i+{\bf t}_0),
    \end{aligned}
    \label{eq:perspective}
\end{equation}
where $\lambda_i$ is a scale factor, and $\bR_0$ and ${\bf t}_0$ are the rotation matrix and translation vector representing the initial pose ${\bf P}_0$. Similarly, we obtain a new 2D location ${\bf u}_{ik}$ of the same 3D point ${\bf p}_i$ under the currently estimated pose ${\bf P}_k$. Then the pose-induced flow $f = {\bf u}_{ik} - {\bf u}_{i0}$, which represents the displacement of 2D reprojection of the same 3D point between the initial pose and the currently estimated pose. We compute 2D flow only for the 3D points on the visible surface of the mesh. Fig.~\ref{fig:correlation_lookup} illustrates the advantages of this strategy.

\subsection{Learning Object Pose From Optical Flow}
\label{sec:learning-pose}

The pose-induced flow relies on the current pose prediction. In principle, the pose can be obtained by a PnP solver based on the current intermediate flow~\cite{PFA}. This strategy, however, is not easy to be stably differentiable during training~\cite{Single-stage_2020_cvpr,wang2021robust}. Instead, we propose using networks to learn the object pose.

We learn a residual pose $\Delta {\bf P}_{k}$ based on the current intermediate flow, which updates the estimated pose iteratively: ${\bf P}_{k} = {\bf P}_{k-1} \otimes \Delta {\bf P}_{k}$. Note that the intermediate flow does not preserve the target's shape, as shown in Fig.~\ref{fig:arch}.

For the supervision of the residual pose, we first encode the residual rotation $\Delta {\bf R}$ into a six-dimensional representation~\cite{rotation_2019_cvpr}, and parameterize the residual translation $\Delta {\bf T}$ as a form of 2D offsets and scaling on the image plane~\cite{DeepIM_2018_eccv}.

\begin{figure}[t]
    \centering
    \setlength\tabcolsep{1pt}
    \begin{tabular}{cc@{\hspace{0.5em}}cc}
    
    \setlength{\fboxsep}{-0.5pt}\setlength{\fboxrule}{0.5pt}\fbox{\includegraphics[width=0.24\linewidth]{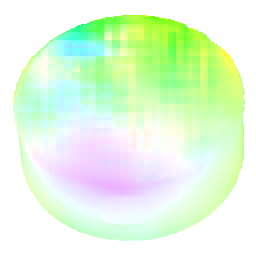}} &
   \setlength{\fboxsep}{-0.5pt}\setlength{\fboxrule}{0.5pt}\fbox{\includegraphics[width=0.24\linewidth]{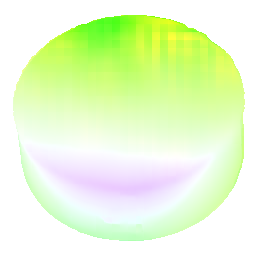}} &
   \setlength{\fboxsep}{-0.5pt}\setlength{\fboxrule}{0.5pt}\fbox{\includegraphics[width=0.24\linewidth]{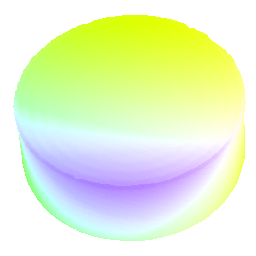}} &
   \setlength{\fboxsep}{-0.5pt}\setlength{\fboxrule}{0.5pt}\fbox{\includegraphics[width=0.24\linewidth]{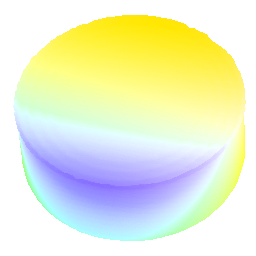}} \\
   \includegraphics[width=0.24\linewidth]{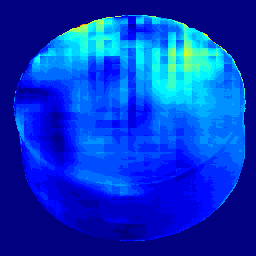} &
  \includegraphics[width=0.24\linewidth]{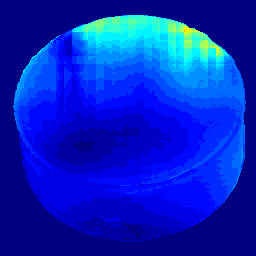} &
   \includegraphics[width=0.24\linewidth]{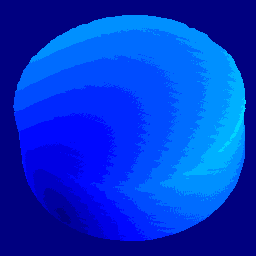} &
   \includegraphics[width=0.24\linewidth]{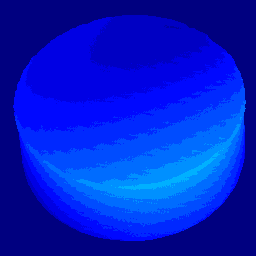} \\
    \setlength{\fboxsep}{-0.5pt}\setlength{\fboxrule}{0.5pt}\fbox{\includegraphics[width=0.24\linewidth]{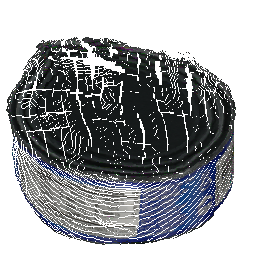}} &
    \setlength{\fboxsep}{-0.5pt}\setlength{\fboxrule}{0.5pt}\fbox{\includegraphics[width=0.24\linewidth]{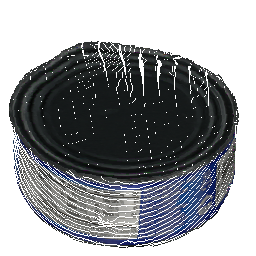}} &
    \setlength{\fboxsep}{-0.5pt}\setlength{\fboxrule}{0.5pt}\fbox{\includegraphics[width=0.24\linewidth]{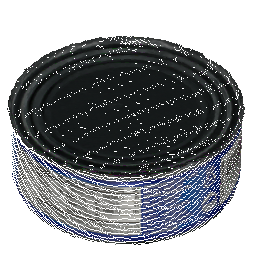}} &
   \setlength{\fboxsep}{-0.5pt}\setlength{\fboxrule}{0.5pt}\fbox{\includegraphics[width=0.24\linewidth]{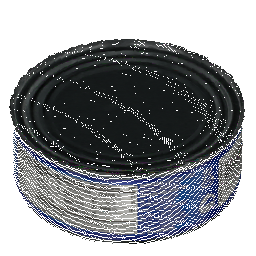}} \\
 
    \multicolumn{2}{c}{\small{(a) The baseline strategy}} & \multicolumn{2}{c}{\small{(b) Our strategy}} \\
\end{tabular}

    \caption{{\bf Comparison of the predicted flow.}
    From top to bottom, we show the predicted flow, the flow error map, and the corresponding warp images for both the baseline PFA and our method. We show the results with 2 and 8 iterations from left to right for each method. The baseline can not preserve the target's shape during the iterations, and our method preserves it after every iteration.
    }
    \label{fig:seq_flow}
\end{figure}

After predicting the current pose ${\bf P}_{k}$, we compute the pose-induced flow based on the initial pose and ${\bf P}_{k}$, as discussed in the previous section. The pose-induced flow, which is used to construct the correlation map for the next iteration, preserves the targets's shape, as shown in Fig.~\ref{fig:seq_flow}.

\subsection{Implementation Details}
\label{sec:implement-details}
Our method works in a render-and-compare manner~\cite{Repose_2021_iccv, DeepIM_2018_eccv, RNNPose_2022_cvpr} to learn the difference between the input image $I_1$ and $I_2$. For $I_1$, we use Pytorch3D~\cite{Pytorch3d_2020_arxiv} to render the target according to the initial pose ${\bf P}_0$, with a fixed image resolution of 256$\times$256. For $I_2$, we crop the region of interest from the raw input image based on ${\bf P}_0$ and resize it to the same resolution as $I_1$.

We use GRU~\cite{GRU, RAFT_2020_eccv} for the recurrent modeling
\begin{equation}
h_{k} = {\rm GRU}(C_k, F_{k-1}, h_{k-1}; {\Theta}),
\end{equation}
where, $C_k$ is the constructed correlation map for the current iteration, $F_{k-1}$ is the intermediate flow from the previous iteration, and $h_k$ and $\Theta$ are the hidden state feature and network parameters of the GRU structure, respectively. 

For the pose regressor, we concatenate the intermediate flow and the hidden state feature of GRU, and use two networks to predict $\Delta {\bf R}$ and $\Delta {\bf T}$, respectively. These two small networks have the same architecture except for the dimension of the final output layer, and consist of three convolutional layers and two fully connected layers for each.

\begin{table*}
    \centering
    \begin{tabular}[]{lcccccccc}
        \toprule
        Dataset             & PoseCNN   & PVNet     & SO-Pose           & DeepIM    & RePose    & RNNPose               & PFA           & {\bf Ours}\\
        \midrule
        LM                  & 63.3      & 86.3      & 96.0              & 88.6      & 96.1      & \underline{97.4}      & 95.8          & {\bf 99.3} \\
        LM-O                & 24.9      & 40.8      & 62.3              & 55.5      & 51.6      & 60.7                  & \underline{65.3}                  & {\bf 66.4}\\
            YCB-V  & 21.3      & -         & 56.8              & 53.6      & 62.1      & \underline{66.4}                           &62.8            & \textbf{70.5}\\

        \bottomrule
\end{tabular}

    \caption{\normalsize{\textbf{Comparison against the state of the art in ADD-0.1d.}
 Our method outperforms the competitors by a large margin.
    }
    }
    \label{tab:compare all dataset}
\end{table*}

We predict both the optical flow and the object pose iteratively. To supervise them in each iteration, we use a simple exponentially weighting strategy~\cite{RAFT_2020_eccv} in our loss
\begin{equation}
   \mathcal{L} = \sum_{k=1}^{N}\gamma^{N-k}(\mathcal{L}^k_{pose} + \alpha \mathcal{L}^k_{flow})
\end{equation}
where $\gamma$ is the exponential weighting factor, and $\alpha$ is the parameter balancing the object pose loss $\mathcal{L}_{pose}$ and the optical flow loss $\mathcal{L}_{flow}$. We use $N = 8$, $\gamma = 0.8$ and $\alpha = 0.1$ in this work.

To compute the object pose loss $\mathcal{L}_{pose}$, we randomly select 1k 3D points from the surface of the object’s 3D mesh, and then calculate the distance between these points transformed by the ground truth pose and the predicted pose, respectively.
For the optical flow loss $\mathcal{L}_{flow}$, we first compute the ground truth flow based on the initial pose and the ground truth pose, by the geometry reasoning as discussed in Section~\ref{sec:shape-constraint}. Then we use the L1 loss to capture the endpoint error between the ground truth flow and the intermediate flow.
We only supervise the pixels within the mask of the rendered target, and discard the pixels under occlusion.

\par

We train our model using AdamW~\cite{AdamW_2019_iclr} optimizer with a batch size of 16, and use an adaptive learning rate scheduler based on One-Cycle~\cite{smith2019super}, starting from 4e-4. We typically train the model for 100k steps. During training, we randomly generate a pseudo initial pose around the ground truth pose of the input image, and render the reference image $I_1$ according to the pseudo initial pose on the fly.

\section{Experiments}

In this section, we evaluate our method systematically. We first introduce our experiment settings and then demonstrate the effectiveness of our method by comparing with the state of the art. Finally, we conduct extensive ablation studies to validate the design of our method. Our source code is publicly available at \url{https://github.com/YangHai-1218/SCFlow}.

\subsection{Experiment Setup}

\noindent \textbf{Datasets.}
We evaluate our method on three challenging datasets, including LINEMOD (``LM'')~\cite{Linemod_2012_accv}, LINEMOD-Occluded (``LM-O'')~\cite{OccLinemod_2015_iccv}, and YCB-V~\cite{PoseCNN_2018_rss}. LINEMOD contains 13 sequences, each containing a single object annotated with accurate ground-truth poses. LINEMOD-Occluded has 8 objects which is a subset of the LM objects. 
Its test set is one of the sequences in LM, which contains all the annotations of the 8 objects in the scene. There is no standard experiment setting on LM and LM-O. Some previous methods~\cite{Gdr-net_2021_cvpr, So-Pose_2022_iccv, RNNPose_2022_cvpr} use different training settings for LM and LM-O, and some methods train a separated model for every single object~\cite{PVNet_2019_cvpr,Repose_2021_iccv}. For consistency and simplicity, we train a single model for all the objects on both LM and LM-O, and for each sequence, we use about 15\% of the RGB images for training, resulting in a total of 2.4k images.
YCB-V is a more challenging dataset containing 21 objects and 130k real images in total, which is captured in cluttered scenes. Besides, we conduct some ablation studies with the BOP synthetic datasets~\cite{bop_2020_eccvw} that include the same objects as those in LM and YCB-V but generated with physically-based rendering (PBR) techniques~\cite{BlenderProc_2019_arxiv}. We train our model only with the real data if
not explicitly stated.

\begin{table}[]
    \centering

\begin{tabular}{lcccc}
    \toprule
    Method      &  Avg.     & MSPD  & VSD   & MSSD \\
    \midrule
    \multicolumn{5}{c}{\textit{YCB-V (Real+PBR)}} \\ 
    \midrule
    {\bf Ours}  & {\bf 0.826}& {\bf 0.860} & \underline{0.778}  &\underline{0.840}\\
    PFA        & 0.795     & 0.844  & 0.743   &0.797    \\
    CIR         & \underline{0.824}     & 0.852 & {\bf 0.783} & 0.835 \\
    CosyPose    & 0.821     & \underline{0.850} & 0.772 & {\bf 0.842} \\
    SurfEmb     & 0.781     & -     &   -   &   -   \\
    \midrule
    \multicolumn{5}{c}{\textit{YCB-V (PBR)}} \\
    \midrule
    {\bf Ours}  & {\bf 0.651}& \underline{0.769}  & {\bf 0.556}   & {\bf 0.626}  \\
    PFA        & 0.615     & 0.739         & 0.521 & 0.585  \\
    SurfEmb     & \underline{0.647}     & {\bf 0.773}   & \underline{0.548} & \underline{0.620} \\
    CosyPose    & 0.574     & 0.653         & 0.516 & 0.554 \\
    \midrule
    \rowcolor{white}
    \multicolumn{5}{c}{\textit{LM-O (PBR)}} \\
    \midrule
    {\bf Ours}  & {\bf 0.682}& \underline{0.842} & {\bf 0.532}      &  {\bf 0.674}\\
    PFA        & \underline{0.674}      & 0.818            & \underline{0.531}  & \underline{0.673}   \\
    CIR         & 0.655     & 0.831             & 0.501 & 0.633 \\
    SurfEmb     & 0.647     & {\bf 0.851}       & 0.497 & 0.640 \\
    CosyPose    & 0.633     & 0.812             & 0.480 & 0.606 \\
    \bottomrule
\end{tabular}
    \caption{{\bf Refinement comparison in BOP metrics.}
We compare our method with the state-of-the-art refinement methods, and our method achieves the best accuracy in different settings in most metrics.
    }
    \label{tab:compare_bop}
\end{table}

\noindent \textbf{Evaluation metrics.}
We use the standard ADD(-S) metric to report the results, which is based on the mean 3D distance between the mesh vertices transformed by the predicted pose and the ground-truth pose, respectively.
We report most of our results in ADD-0.1d, which is the percentage of correct predictions with a distance error less than 10\% of the mesh diameter. In some ablation studies, we report ADD-0.05d, which uses a threshold of 5\% of the model diameter.
Some recent methods~\cite{cosypose_2020_eccv, surfemb, coupled_2022_cvpr} only report their results in BOP metrics~\cite{bop_2020_eccvw}. For comparing with them, we report some of our results in BOP metrics which include the Visible Surface Discrepancy (VSD), the Maximum Symmetry-aware Surface Distance (MSSD), and the Maximum Symmetry-aware Projection Distance (MSPD).
We refer readers to ~\cite{bop_2020_eccvw} for the detailed metric definition.

\subsection{Comparison to the State of the Art}
We compare our method with most state-of-the-art methods, including PoseCNN~\cite{PoseCNN_2018_rss}, PVNet~\cite{PVNet_2019_cvpr}, SO-Pose~\cite{So-Pose_2022_iccv}, DeepIM~\cite{DeepIM_2018_eccv}, RePose~\cite{Repose_2021_iccv}, RNNPose~\cite{RNNPose_2022_cvpr}, and PFA~\cite{PFA}.
We use the results of PoseCNN as our pose initialization by default.
For PFA, we use its official code which relies on online rendering and has only one view for correspondence generation, producing slightly better results than that in its paper.
Nevertheless, our method outperforms most of them by a large margin, as shown in Table~\ref{tab:compare all dataset}.

Furthermore, we compare with the recent refinement methods CIR~\cite{coupled_2022_cvpr}, CosyPose~\cite{cosypose_2020_eccv}, and SurfEmb~\cite{surfemb}, which only have results in BOP metrics. Since all of them use the first stage of CosyPose as their pose initialization, we use the same initialization in this experiment for a fair comparison.
We report the results on YCB-V and LM-O, and on LM-O we only report the results with PBR training images following the standard BOP setting.
As shown in Table~\ref{tab:compare_bop}, our method achieves the best accuracy in most metrics.

\noindent \textbf{Running time analysis.}
We evaluate our method on a workstation with an NVIDIA RTX-3090 GPU and an Intel-Xeon CPU with 12 2.1GHz cores. We run most of the recent refinement methods on the same workstation. Since different refinement methods use very different strategies for pose initialization, we only report the running time of pose refining, as shown in Table~\ref{tab:profile}. Our method takes only 17ms on average to process an object instance and is much more efficient than most refinement methods. Thanks to the removal of the need for RANSAC-PnP, our method is more than twice faster than PFA.

\begin{table}[]
    \centering
    \setlength\tabcolsep{2pt}
    
\begin{tabular}{c@{\hspace{2em}}c@{\hspace{1em}}c@{\hspace{1em}}c@{\hspace{1em}}c@{\hspace{1em}}c@{\hspace{1em}}c}
    \toprule
    Method              & CIR      & CosyPose      & SurfEmb       & PFA       & {\bf Ours} \\
    \midrule
    Timing             &  11k            &   20          & 1k          &  37       & {\bf 17}     \\
    \bottomrule
\end{tabular}
    \caption{{\bf Efficiency comparison.}
    We run all the methods on the same workstation, and report the running time in milliseconds in processing an image containing one object instance. Our method is the most efficient one among all the competitors.
    }
    \label{tab:profile}
\end{table}

\begin{table}
    \centering
    \begin{tabular}{lccc}
    \toprule
     Method    & LM        & LM-O       & YCB-V\\
    \midrule
    {WDR}        & 60.2      & 37.9      & 27.5  \\
       w/ PFA  & {\bf 95.8}      & 65.3      & 62.8\\
       w/ {\bf Ours} & {\bf 95.8}      & {\bf 71.1}      & {\bf 65.5}\\
    \midrule
    PoseCNN       & 63.3      & 24.9      & 21.3  \\
       w/ PFA   & 99.2       & 60.5       & 61.9  \\
        w/ {\bf Ours} & {\bf 99.3}      & {\bf 66.4}     & {\bf 70.5}  \\
    \bottomrule
\end{tabular}

    \caption{\textbf{Comparison with different pose initialization.}
We compare our refinement method with the baseline PFA with different pose initialization, including WDR and PoseCNN. Our method consistently outperforms PFA in each setting.
    }
    \label{tab:different init}
 \end{table}

\begin{figure}
    \centering
    \begin{tabular}{cc}
        \includegraphics[width=0.45\linewidth]{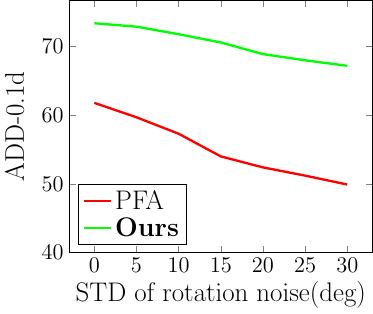} &
        \includegraphics[width=0.45\linewidth]{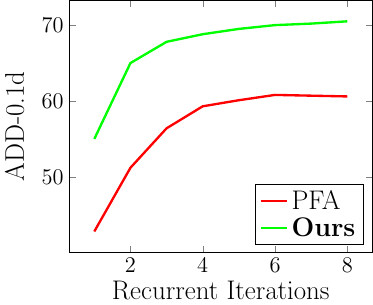} \\
        \small{(a) Robustness} & \small{(b) Effectiveness} \\
    \end{tabular}\
    \caption{{\bf Ablation study on YCB-V.}
    {\bf (a)} We compare our method and PFA with the results of PoseCNN as pose initialization with random pose errors in different levels.
    Our method is much more robust than PFA, especially in scenarios with heavy initialization noise.
    {\bf (b)} We evaluate the methods with different recurrent iterations during inference, and our method outperforms PFA after only 2 iterations, and improves further with more iterations.
    }
    \label{fig:robustness}
\end{figure}

\begin{table}
    \centering
    \begin{tabular}{ccc@{\hspace{2em}}cc}
    \toprule
    Shape-constraint & Pose & ADD & ADD \\
    lookup & regressor & 0.05d & 0.1d\\
    \midrule
    -                    & -                  & 33.5     & 61.9  \\ 
    -                    & \cmark                            & 34.0      &  63.6     \\ 
    \cmark            & -                   & 46.1      & 67.6  \\ 
    \cmark            & \cmark                   &  {\bf 50.4}     & {\bf 70.5} \\ 
    \bottomrule 
\end{tabular}
    \caption{\textbf{Ablation study of different components on YCB-V.}
We evaluate the two key components of our method, including the shape-constraint lookup operation and the pose regressor. For the notation ``-'' for them, we use the standard constraint-free lookup operation based on the intermediate flow, and the RANSAC-PnP, respectively. The first row is the baseline PFA. Our shape-constraint lookup boosts the performance, and the pose regressor increases the performance further.
    }
    \label{tab:ablation_study}
\end{table}

\subsection{Ablation Study}
\noindent \textbf{Robustness to different pose initialization.}
Our pose refinement method is general and can be used with most pose methods with their results as pose initialization. To verify the generalization ability of our method, we evaluate our method with pose initialization from the results of WDR~\cite{WDR_2021_cvpr} and PoseCNN~\cite{PoseCNN_2018_rss}, respectively.
For WDR, we obtain its results trained only on the synthetic data, similar to PFA~\cite{PFA}. For PoseCNN, we use its pre-generated results, which was trained on real images. As shown in Table~\ref{tab:different init}, our method improves the results of pose initialization significantly and outperforms PFA in most settings.

Furthermore, we study the robustness of our method with the results of PoseCNN as pose initialization with random pose errors in different levels, as shown in Fig.~\ref{fig:robustness}(a). Our method is much more robust than PFA. Especially, the performance drop of PFA with heavy noise can be nearly 11.9\% in ADD-0.1d, and our method's accuracy only decreases by about 6.2\% in the same condition and is still higher than that of PFA obtained with little initial noise.

\noindent \textbf{Effect of different number of iterations.}
We evaluate our method with different number of iterations during inference. As shown in Fig.~\ref{fig:robustness}(b), our method performs on par with PFA after the very first iteration, and outperforms it significantly after only 2 iterations. Our method improves further with more iterations, and outperforms PFA by over 11.5\% in the end.

\noindent \textbf{Evaluation of different components.}
We study the effect of different components of our method, including the shape-constraint lookup operation guided by the pose-induced flow, and the pose regressor based on the intermediate flow,  as shown in Table~\ref{tab:ablation_study}.
The first row is the baseline method PFA, which does not have any constraints in the correlation map construction and relies on RANSAC-PnP. Our shape-constraint lookup operation boosts the performance, demonstrating the effectiveness of embedding targets' 3D shape. The RANSAC-PnP, even equipped with the shape-constraint lookup during the recurrent optimization, still suffers in producing accurate pose results, which is caused by the surrogate loss that does not directly reflect the final object pose. By contrast, our pose regressor is end-to-end trainable, which does not suffer from this problem and can benefit from simultaneously optimizing the optical flow and object pose.

\noindent \textbf{Evaluation with different training data.}
To study the effect of different training data, we report the results of our method trained with four data settings, including pure PBR images, PBR images with additional 20 real images for each object (``PBR+20''), pure real images, and a mixture of all PBR images and real images. 
As shown in Table~\ref{tab:eval_on_different_data}, more data generally results in more accurate pose estimates. While we report most results of our method trained only on the real images to be consistent with other methods.
On the other hand, we find that, on LM, the models trained with only real images perform even better than those trained with a mixture of the real and PBR images, which we believe is caused by the distribution difference between the PBR and real data on LM.
Note that the results of PFA are different from that in Table~\ref{tab:compare all dataset} since here we use the same results of PoseCNN as pose initialization for both methods for fair comparison.
Nevertheless, our method consistently outperforms PFA in all different settings.

\begin{table}
    \centering
    \small
    \begin{tabular}{c|cc|cc|cc}
       \toprule
       Dataset     & \multicolumn{2}{c|}{LM}       & \multicolumn{2}{c|}{LM-O}  & \multicolumn{2}{c}{YCB-V}\\ 
                   & 0.05d  & 0.1d                & 0.05d     & 0.1d          & 0.05d  & 0.1d  \\ 
       \midrule
                                        & 65.5   & 95.0                &  27.6     &  50.9         & 8.8   & 28.9    \\ 
        \multirow{-2}{*}{PBR}           & {\bf 72.5}   & {\bf 96.8}                & {\bf 28.9}      & {\bf 52.9}          & {\bf 10.9}   & {\bf 36.5}  \\ 
        \cmidrule{2-7}
        \multirow{2}{*}{PBR+20}         & 77.3          & 97.7                &  28.2     &  54.5         & 24.0   & 49.6  \\ 
                                        & {\bf 81.5}    & {\bf 98.5}            & {\bf 36.2}      & {\bf 63.3}          & {\bf 33.2}   & {\bf 61.9}  \\ 
        \cmidrule{2-7}
                   & 89.5   & 99.2                &  30.6     &  60.5         & 33.5   & 61.9 \\ 
        \multirow{-2}{*}{Real}       & \textbf{92.9}&\textbf{99.3}  & {\bf 39.3}      & {\bf 66.4}          & {\bf 50.4}   & {\bf 70.5}  \\
        \cmidrule{2-7}
        \multirow{2}{*}{Mixed}       & 77.3   & 97.7                &  37.7     &  64.5         & 38.0   & 61.6  \\ 
               & {\bf 90.9}   & \textbf{99.3}    &   \textbf{44.6} &\textbf{67.0} & {\bf 51.2} & {\bf 73.2}  \\                        
       \bottomrule
\end{tabular}
    \caption{\textbf{Comparison with different training data.}
    We compare our method with the baseline PFA in four different data settings. In each setting, the {\bf first row} is PFA, and the {\bf second row} is ours. Our method outperforms PFA in all settings.
    }
    \label{tab:eval_on_different_data}
\end{table}

\begin{figure}
    \centering
    \begin{tabular}{cc}
        \includegraphics[width=0.45\linewidth]{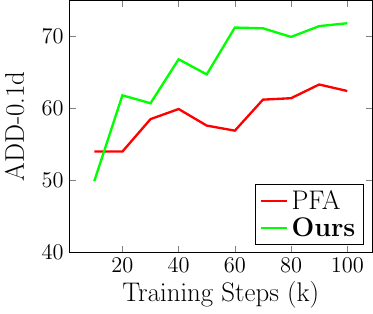} & 
       \includegraphics[width=0.45\linewidth]{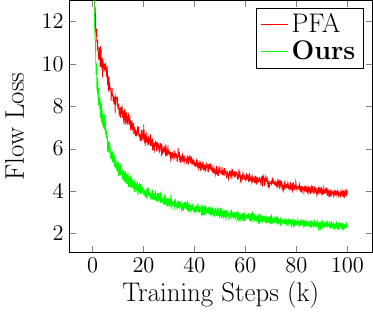}  \\
    \end{tabular}
    \caption{{\bf Training analysis on YCB-V.}
We report the pose accuracy and flow loss during training for both the baseline PFA and our method. Our method performs equally well as the fully-trained PFA after only 20k training steps, and outperforms it significantly with more training steps.
    }
    \label{fig:training_acc}
\end{figure}

\begin{figure}
    \centering
    \begin{tabular}{cc}
        \includegraphics[width=0.45\linewidth]{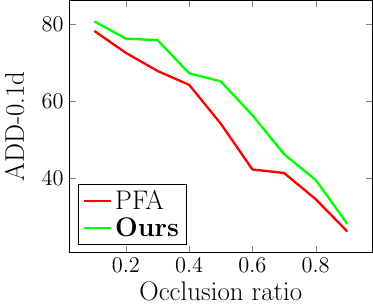} &
        \includegraphics[width=0.45\linewidth]{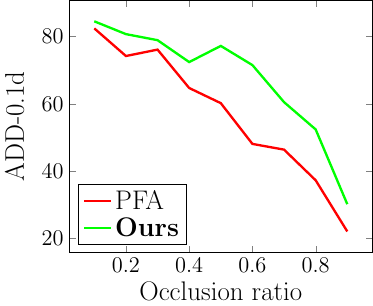}\\
        {\small (a) LM-O}   & {\small (b) YCB-V}\\
    \end{tabular}
    \caption{{\bf Performance with different occlusion ratios.}
    Our method consistently outperforms the baseline PFA in different occlusion ratios, demonstrating the effectiveness of our shape-constraint strategy.
    }
    \label{fig:compare_occlusion}
\end{figure}

\begin{figure*}
   \centering
   \setlength\tabcolsep{1pt}
   \begin{tabular}{cccccccc}
  \includegraphics[width=0.12\linewidth]{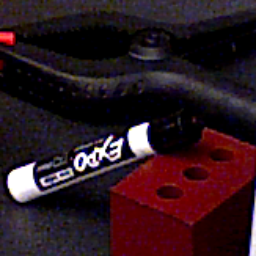}&
  {\setlength{\fboxsep}{-0.5pt}\setlength{\fboxrule}{0.5pt}\fbox{\includegraphics[width=0.12\linewidth]{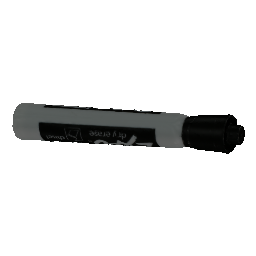}}}&
  \setlength{\fboxsep}{-0.5pt}\setlength{\fboxrule}{0.5pt}\fbox{\includegraphics[width=0.12\linewidth]{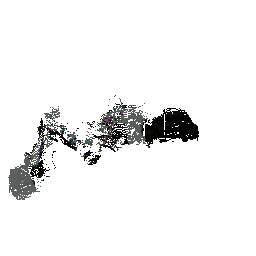}}&
  \setlength{\fboxsep}{-0.5pt}\setlength{\fboxrule}{0.5pt}\fbox{\includegraphics[width=0.12\linewidth]{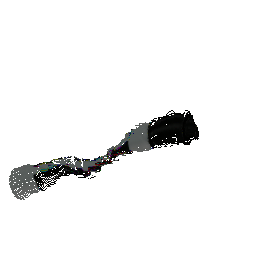}}&
  \includegraphics[width=0.12\linewidth]{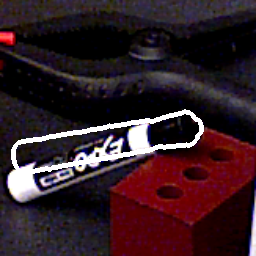}&
  \setlength{\fboxsep}{-0.5pt}\setlength{\fboxrule}{0.5pt}\fbox{\includegraphics[width=0.12\linewidth]{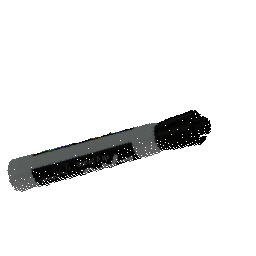}}&
  \setlength{\fboxsep}{-0.5pt}\setlength{\fboxrule}{0.5pt}\fbox{\includegraphics[width=0.12\linewidth]{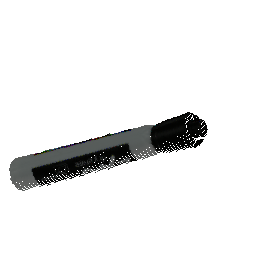}}&
  \includegraphics[width=0.12\linewidth]{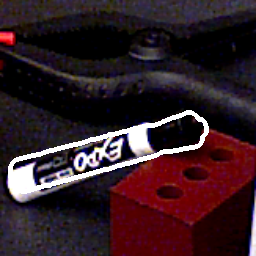}\\
  
  \includegraphics[width=0.12\linewidth]{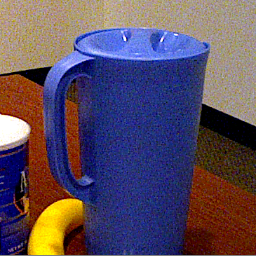} &
  {\setlength{\fboxsep}{-0.5pt}\setlength{\fboxrule}{0.5pt}\fbox{\includegraphics[width=0.12\linewidth]{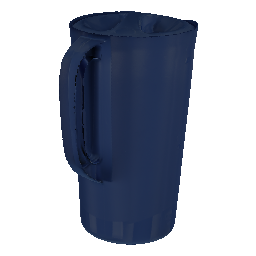}}} &
  \setlength{\fboxsep}{-0.5pt}\setlength{\fboxrule}{0.5pt}\fbox{\includegraphics[width=0.12\linewidth]{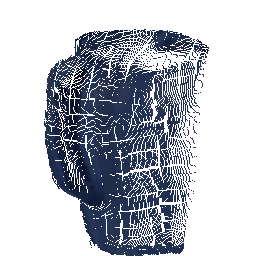}}&
  \setlength{\fboxsep}{-0.5pt}\setlength{\fboxrule}{0.5pt}\fbox{\includegraphics[width=0.12\linewidth]{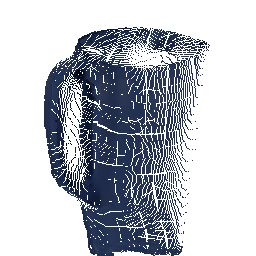}}&
  \includegraphics[width=0.12\linewidth]{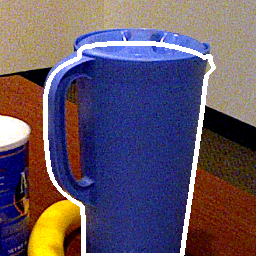}&
  \setlength{\fboxsep}{-0.5pt}\setlength{\fboxrule}{0.5pt}\fbox{\includegraphics[width=0.12\linewidth]{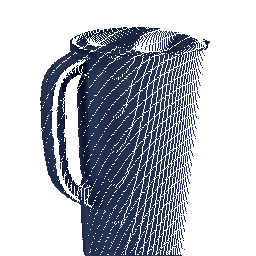}}&
  \setlength{\fboxsep}{-0.5pt}\setlength{\fboxrule}{0.5pt}\fbox{\includegraphics[width=0.12\linewidth]{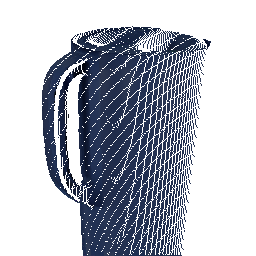}}&
  \includegraphics[width=0.12\linewidth]{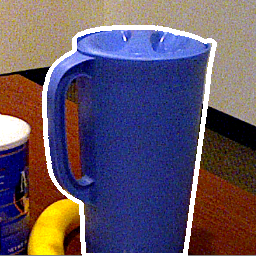}\\

  \includegraphics[width=0.12\linewidth]{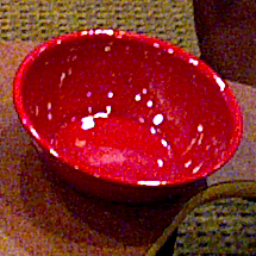} &
  \setlength{\fboxsep}{-0.5pt}\setlength{\fboxrule}{0.5pt}\fbox{\includegraphics[width=0.12\linewidth]{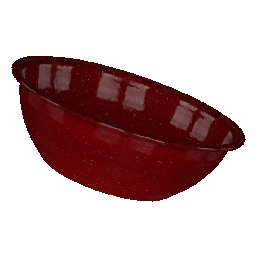}}&
  \setlength{\fboxsep}{-0.5pt}\setlength{\fboxrule}{0.5pt}\fbox{\includegraphics[width=0.12\linewidth]{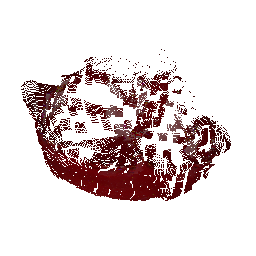} } & 
  \setlength{\fboxsep}{-0.5pt}\setlength{\fboxrule}{0.5pt}\fbox{\includegraphics[width=0.12\linewidth]{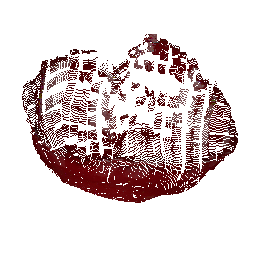} } &
  \includegraphics[width=0.12\linewidth]{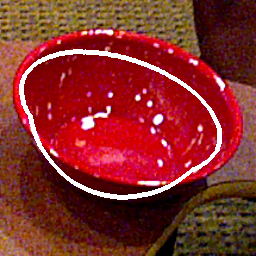} &
  \setlength{\fboxsep}{-0.5pt}\setlength{\fboxrule}{0.5pt}\fbox{\includegraphics[width=0.12\linewidth]{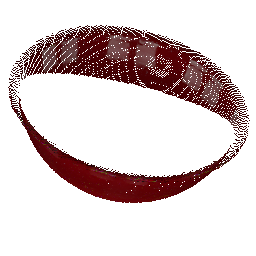} } &
  \setlength{\fboxsep}{-0.5pt}\setlength{\fboxrule}{0.5pt}\fbox{\includegraphics[width=0.12\linewidth]{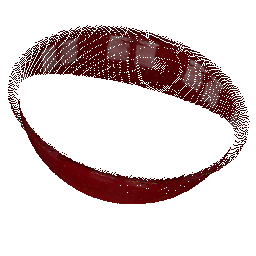} } &
  \includegraphics[width=0.12\linewidth]{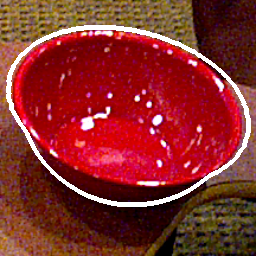} \\

  \includegraphics[width=0.12\linewidth]{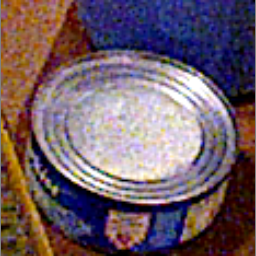} &
  \setlength{\fboxsep}{-0.5pt}\setlength{\fboxrule}{0.5pt}\fbox{\includegraphics[width=0.12\linewidth]{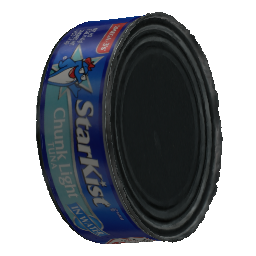}}&
  \setlength{\fboxsep}{-0.5pt}\setlength{\fboxrule}{0.5pt}\fbox{\includegraphics[width=0.12\linewidth]{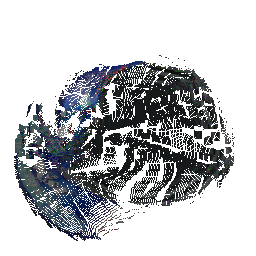} } & 
  \setlength{\fboxsep}{-0.5pt}\setlength{\fboxrule}{0.5pt}\fbox{\includegraphics[width=0.12\linewidth]{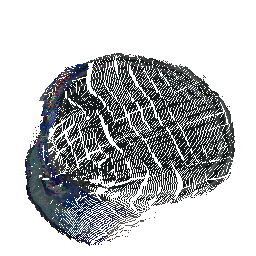} } &
  \includegraphics[width=0.12\linewidth]{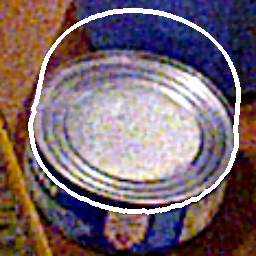} &
  \setlength{\fboxsep}{-0.5pt}\setlength{\fboxrule}{0.5pt}\fbox{\includegraphics[width=0.12\linewidth]{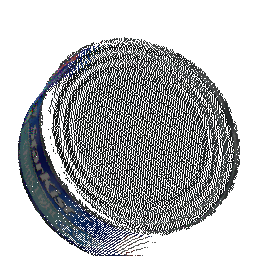} } &
  \setlength{\fboxsep}{-0.5pt}\setlength{\fboxrule}{0.5pt}\fbox{\includegraphics[width=0.12\linewidth]{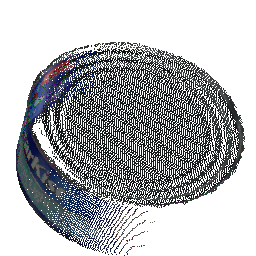} } &
  \includegraphics[width=0.12\linewidth]{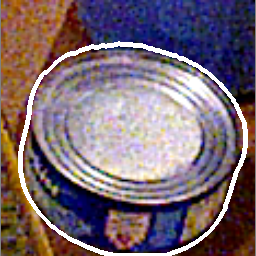} \\

  \includegraphics[width=0.12\linewidth]{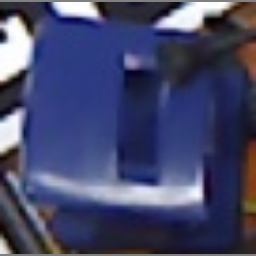} &
  \setlength{\fboxsep}{-0.5pt}\setlength{\fboxrule}{0.5pt}\fbox{\includegraphics[width=0.12\linewidth]{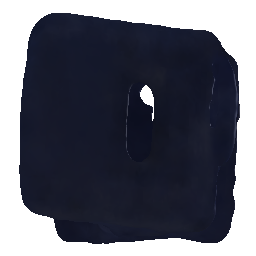}}&
  \setlength{\fboxsep}{-0.5pt}\setlength{\fboxrule}{0.5pt}\fbox{\includegraphics[width=0.12\linewidth]{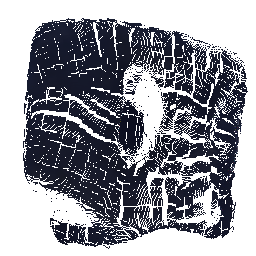} } & 
  \setlength{\fboxsep}{-0.5pt}\setlength{\fboxrule}{0.5pt}\fbox{\includegraphics[width=0.12\linewidth]{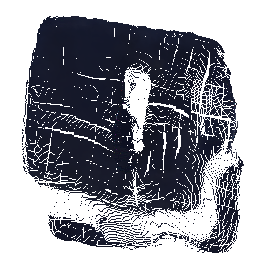} } &
  \includegraphics[width=0.12\linewidth]{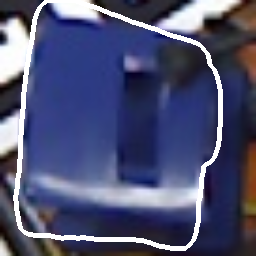} &
  \setlength{\fboxsep}{-0.5pt}\setlength{\fboxrule}{0.5pt}\fbox{\includegraphics[width=0.12\linewidth]{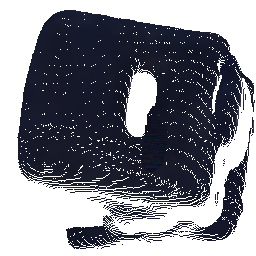} } &
  \setlength{\fboxsep}{-0.5pt}\setlength{\fboxrule}{0.5pt}\fbox{\includegraphics[width=0.12\linewidth]{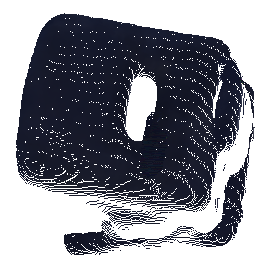} } &
  \includegraphics[width=0.12\linewidth]{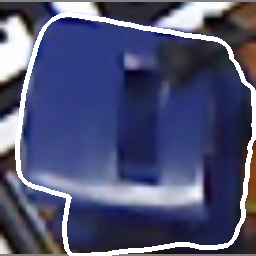} \\
    
  \includegraphics[width=0.12\linewidth]{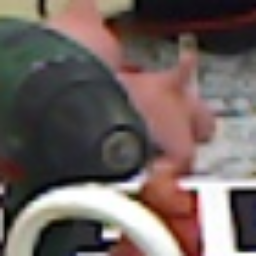} &
  \setlength{\fboxsep}{-0.5pt}\setlength{\fboxrule}{0.5pt}\fbox{\includegraphics[width=0.12\linewidth]{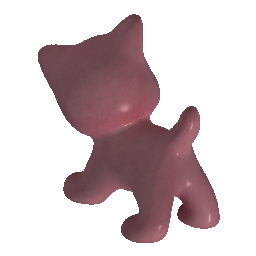}}&
  \setlength{\fboxsep}{-0.5pt}\setlength{\fboxrule}{0.5pt}\fbox{\includegraphics[width=0.12\linewidth]{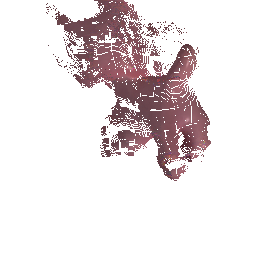} } & 
  \setlength{\fboxsep}{-0.5pt}\setlength{\fboxrule}{0.5pt}\fbox{\includegraphics[width=0.12\linewidth]{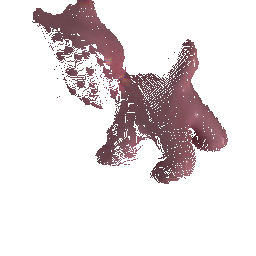} } &
  \includegraphics[width=0.12\linewidth]{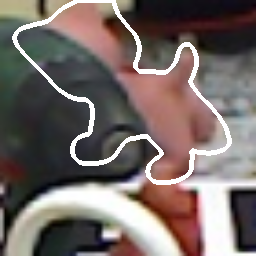} &
  \setlength{\fboxsep}{-0.5pt}\setlength{\fboxrule}{0.5pt}\fbox{\includegraphics[width=0.12\linewidth]{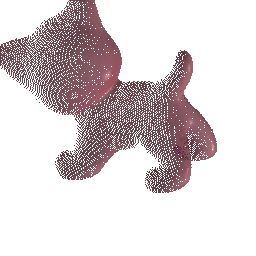} } &
  \setlength{\fboxsep}{-0.5pt}\setlength{\fboxrule}{0.5pt}\fbox{\includegraphics[width=0.12\linewidth]{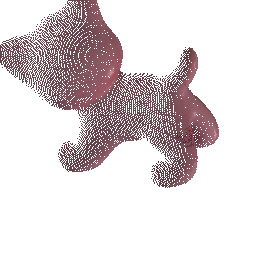} } &
  \includegraphics[width=0.12\linewidth]{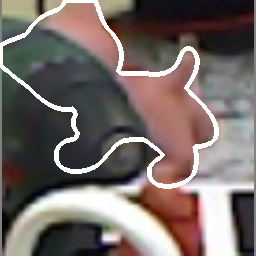} \\
  
  \small{(a) Target}  &  \small{(b) Initialization}    & \multicolumn{3}{c}{{\cellcolor{red!10} \small{(c) PFA}}}  &  \multicolumn{3}{c}{\cellcolor{green!10} \small{{(d) Ours}}} \\
\end{tabular}
   \caption{\textbf{Qualitative results.} 
   We visualize the initialization by rendering the target according to the initial pose. For both PFA and our method, we show the flow wrap results with 2 and 8 iterations from left to right, respectively. PFA can not preserve the target's shape even after 8 iterations. By contrast, our method preserves the shape after every iteration and produces more accurate pose results by the end.
    }
   \label{fig:visualization}
\end{figure*}

\noindent \textbf{Training analysis.}
We study the properties of our method during training. 
We report the pose accuracy and flow loss for both the baseline PFA and our method in different steps during training, as shown in Fig.~\ref{fig:training_acc}. 
At the beginning, our method performs less well than PFA. However, after only 20k training steps, our method outperforms PFA with the same training steps and performs equally well as the fully-trained PFA. Furthermore, our method produces much better results than PFA in both pose accuracy and flow loss with more training steps.

\noindent \textbf{Performance with different occlusion ratios.}
We compare our method with the baseline PFA in scenarios with different occlusion ratios. As shown in Fig.~\ref{fig:compare_occlusion}, although the performance of both methods decreases with the increase of occlusion rations, our method is more robust than PFA and outperforms it in every setting of different occlusion ratio, either on LM-O or YCB-V, thanks to our shape-constraint design, which implicitly embeds the target's 3D shape information into our model and is more robust to occlusions. We show some qualitative results in Fig.~\ref{fig:visualization}.

\section{Conclusion}
We have introduced a shape-constraint recurrent matching framework for 6D object pose estimation. We have first analyzed the weaknesses of the standard optical flow networks and introduced a new matching framework that contains only all the 2D reprojections of the target's 3D shape in constructing the correlation map, which reduces the matching space significantly.
Furthermore, we have proposed learning the object pose based on the current estimated flow and simultaneously optimizing the object pose and optical flow in an end-to-end recurrent manner. We have demonstrated the advantages of our method with extensive evaluation on three challenging 6D object pose datasets. It outperforms the state of the art significantly, and converges much more quickly.

\vspace{0.2em}
{\small
{\noindent \bf Acknowledgments.}
This work was supported by the 111 Project of China under Grant B08038, the Fundamental Research Funds for the Central Universities under Grant JBF220101, and the Youth Innovation Team of Shaanxi Universities.
}

\section{Appendix}
We show the detailed results of each object on LINEMOD, Occluded-LINEMOD, and YCB-V dataset, in Table~\ref{tab:lm_add_005d}, ~\ref{tab:OccLinemod compare}, and ~\ref{tab:ycbv compare}. We report the numbers in ADD-0.1d and denote the symmetry objects with * in the tables.


\begin{table*}
    \centering

\begin{tabular}{l|ccccccc}
    \toprule
    Method          & PVNet     & DeepIM      & CDPN    & {\bf Ours$\dagger$}  &  {\bf Ours}  & {\bf Ours$\dagger$}  &  {\bf Ours} \\
    \midrule
    Training Data   & Real+Syn  & Real+Syn  & Real+Syn  & Real      & Real      & Real+PBR             & Real+PBR       \\  
    \midrule
    ape             & 43.6      & 77.0      & 64.4      & 94.3      & 95.1      & 95.2     & {\bf 96.1}        \\
    benchvise       & 99.9      & 97.5      & 97.8      & {\bf 100.0}     & {\bf 100.0}     & {\bf 100.0}    & {\bf 100.0}       \\ 
    cam             & 86.9      & 93.5      & 91.7      & 95.0      & 99.6      & 95.2     & {\bf 99.5}        \\
    can             & 95.5      & 96.5      & 95.9      & 97.0      & {\bf 99.9}      & 97.1     & {\bf 99.9}         \\
    cat             & 79.3      & 82.1      & 83.8      & 99.5      & 99.4      & 99.5     & {\bf 99.7}         \\
    driller         & 96.4      & 95.0      & 96.2      & 97.0      & {\bf 100.0}     & 97.2     & {\bf 100.0}        \\
    duck            & 52.6      & 77.7      & 66.8      & 92.4      & 91.9      & 94.0     & {\bf 94.1}        \\
    eggbox*         & 99.2      & 97.1      & 99.7      & 95.4      & 99.9      & 95.0     & {\bf 100.0}       \\
    glue*           & 95.7      & 99.4      & 99.6      & 99.1      & {\bf 99.9}      & 98.8     & 99.8          \\
    holepuncher     & 81.9      & 52.8      & 85.8      & 94.7      & {\bf 97.6}      & 95.3     & 97.5           \\
    iron            & 98.9      & 98.3      & 97.9      & 99.9      & 99.9      & 99.9     & {\bf 100.0}       \\
    lamp            & 99.3      & 97.5      & 97.9      & 98.6      & {\bf 99.8}      & 98.4     & {\bf 99.8}         \\
    phone           & 92.4      & 87.7      & 90.8      & 93.1      & 99.3      & 93.7     & {\bf 100.0}        \\
    \midrule
    Avg.            & 86.3      & 88.6      & 89.9      & 96.6      & 98.6      & 96.9     & {\bf 99.3}         \\
    \bottomrule
\end{tabular}
    \caption{{\bf Results on LINEMOD.}
    We compare our method with PVNet~\cite{PVNet_2019_cvpr}, DeepIM~\cite{DeepIM_2018_eccv}, and CDPN~\cite{CDPN_2019_iccv}. 
    $\dagger$ denotes the results obtained by the initial pose from WDR.
    }
    \label{tab:lm_add_005d}
    \label{tab:my_label}
\end{table*}

\begin{table*}
    \centering
    \setlength\tabcolsep{5pt}
    \begin{tabular}{lccccccc}
   \toprule
   Method         & RePose    & DeepIM      & RNNPose       & {\bf Ours}    & {\bf Ours$\dagger$}   & {\bf Ours}         & {\bf Ours$\dagger$}      \\
   \midrule
   Training Data  & Real+Syn  & Real+Syn    & Real+Syn      & Real           & Real          & Real+PBR                   & Real+PBR         \\    
   \midrule
   ape            & 31.1      & 59.2        & 37.2          & 50.7           & 55.6           & 51.3                & {\bf 57.4} \\
   can            & 80.0      & 63.5        & 88.1          & 86.9           & 91.2           & 89.5                & {\bf 93.4}\\
   cat            & 25.6      & 26.2        & 29.2          & 51.6           & 53.2           & 53.6                & {\bf 53.9}\\
   driller        & 73.1      & 55.6        & 88.1          & 93.0           & 94.3           & {\bf 95.0}          & 94.0\\
   duck           & 43.0      & 52.4        & 49.2          & 55.3           & 60.1           & 55.9                & {\bf 61.2}\\
   eggbox*        & 51.7      & 63.0        & 43.2          & 38.5           & 60.4           & 39.8                & {\bf 67.4}\\
   glue*          & 54.3      & 71.7        & 63.8          & 79.2           & 81.0           & {\bf 82.5}          & 81.7\\
   hole.          & 53.6      & 52.5        & 62.8          & 76.0           & 72.8           & 68.5                & \textbf{76.2}\\
   \midrule
   Avg.           & 51.6      & 55.5        & 60.7          & 66.4           & 71.1            & 67.0                & \textbf{73.1} \\ 
   \bottomrule

\end{tabular}
    \caption{\textbf{Results on Occluded-LINEMOD.}
The results of our method are significantly better than RePose~\cite{Repose_2021_iccv}, DeepIM~\cite{DeepIM_2018_eccv}, and RNNPose~\cite{RNNPose_2022_cvpr}.
$\dagger$ denotes the results obtained by the initial pose from WDR.
    }
    \label{tab:OccLinemod compare}
\end{table*}

\begin{table*}
    \centering
    \setlength\tabcolsep{4pt}

\begin{tabular}{lccccccc}
  \toprule
  Method                      & PoseCNN  & SegDriven    &GDR-Net        &\textbf{Ours$\dagger$} & \textbf{Ours} & \textbf{Ours$\dagger$}   & \textbf{Ours} \\
  \midrule
  Training Data               & Real      & Real+Syn    & Real+PBR      & Real      & Real      & Real+PBR          & Real+PBR \\
  \midrule
  002\_master\_chef\_can       &  3.6     & 33.0        & 41.5          & {\bf 73.3}& 55.3        & 68.3            & 65.3\\
  003\_cracker\_box            & 25.1     & 44.6        & 83.2          & 93.8      & 94.7        & 94.7            & {\bf 99.6}\\
  004\_sugar\_box              & 40.3     & 75.6        & 91.5          & 99.5      & 99.5        & 99.5            & {\bf 99.7}\\
  005\_tomato\_soup\_can       & 25.5     & 40.8        & 65.9          & {\bf 82.7}& 72.7        & 81.8            & 74.2\\
  006\_mustard\_bottle         & 61.9     & 70.6        & 90.2          & 99.3      & 99.3        & 99.3            & {\bf 100.0}\\
  007\_tuna\_fish\_can         & 11.4     & 18.1        & 44.2          & 70.0      & 63.3        & {\bf 73.0}      & 68.0\\
  008\_pudding\_box            & 14.5     & 12.2        & 2.8           & 70.7      & {\bf 82.7}  & 81.3            & 80.0\\
  009\_gelatin\_box            & 12.1     & 59.4        & 61.7          & 92.0      & 89.3        & {\bf 98.7}      & 89.3\\
  010\_potted\_meat\_can       & 18.9     & 33.3        & 64.9          & 68.0      & {\bf 69.3}  & 67.1            & 67.6\\
  011\_banana                  & 30.3     & 16.6        & 64.1          & 80.0      & 78.0        & {\bf 84.0}      & {\bf 84.0}\\
  019\_pitcher\_base           & 15.6     & 90.0        & 99.0          & 95.1      & 99.1        & 95.1            & {\bf 100.0}\\
  021\_bleach\_cleanser        & 21.2     & 70.9        & 73.8          & 56.3      & 74.0        & 54.3            & {\bf 82.3}\\
  024\_bowl*                   & 12.1     & 30.5        & 37.7          & 26.0      & {\bf 50.0}  & 24.7            & {\bf 50.0}\\
  025\_mug                     &  5.2     & 40.7        & 61.5          & 52.7      & 54.7        & {\bf 70.0}      & 64.0\\
  035\_power\_drill            & 29.9     & 63.5        & 78.5          & 97.3      & {\bf 98.0}  & 95.7            & {\bf 98.0}\\
  036\_wood\_block*            & 10.7     & 27.7        & 59.5          & 0.0       & 74.7        & 0.0             & {\bf 76.0}\\
  037\_scissors                &  2.2     & 17.1        & 3.9           & 10.2      & {\bf 17.3}  & 6.7             & 16.0\\
  040\_large\_marker           & 3.4      & 4.8         & 7.4           & 2.7       & {\bf 10.0}  & 2.0             &  8.7\\
  051\_large\_clamp*           & 28.5     & 25.6        & 69.8          & 82.7      & 79.3        & {\bf 83.3}      & 79.3\\
  052\_extra\_large\_clamp*    & 19.6     &  8.8        & 90.0          & 52.0      & 56.7        & 53.3            & {\bf 56.7}\\
  061\_foam\_brick*            & 54.5     & 34.7        & 71.9          & 72.0      & 63.0        & 72.0            & {\bf 77.3}\\
  \midrule
  MEAN                         & 21.3     & 39.0        & 60.1          & 65.5      & 70.5        & 66.9            & {\bf 73.2}\\
  \bottomrule
\end{tabular}

    \caption{{\bf Results on YCB-V.}
We compare to PoseCNN~\cite{PoseCNN_2018_rss}, SegDriven~\cite{Segdriven_2019_cvpr}, and GDR-Net~\cite{Gdr-net_2021_cvpr}.
$\dagger$ denotes the results obtained by the initial pose from WDR.
    }
    \label{tab:ycbv compare}
\end{table*}

{\small
\bibliographystyle{ieee_fullname}
\bibliography{egbib}
}

\end{document}